# IMPROVING THE QUALITY OF GUJARATI-HINDI MACHINE TRANSLATION THROUGH PART-OF-SPEECH TAGGING AND STEMMER ASSISTED TRANSLITERATION

Juhi Ameta[1], Nisheeth Joshi[2] and Iti Mathur[3]

[1]Department of Computer Engineering, Cummins College of Engineering for Women, Pune, Maharashtra, India
[2,3]Department of Computer Science, Apaji Institute, Banasthali University, Rajasthan, India

[1]juhiameta.trivedi@gmail.com
[2]nisheeth.joshi@rediffmail.com
[3]mathur_iti@rediffmail.com

*ABSTRACT*

*Machine Translation for Indian languages is an emerging research area. Transliteration is one such module that we design while designing a translation system. Transliteration means mapping of source language text into the target language. Simple mapping decreases the efficiency of overall translation system. We propose the use of stemming and part-of-speech tagging for transliteration. The effectiveness of translation can be improved if we use part-of-speech tagging and stemming assisted transliteration. We have shown that much of the content in Gujarati gets transliterated while being processed for translation to Hindi language.*

## 1. INTRODUCTION

Transliteration is a process that transliterates or rather maps the source content to the target content. While we design a translation model, transliteration proves to be an effective means for those words which are multilingual or which are not present in the training corpus. For a highly inflectional Indian language like Gujarati, naive transliteration i.e. direct transliteration without any rules or constraints, does not prove to be very effective. The main reason behind this is that suffixes get attached to the root words while forming a sentence.

We propose the use of stemming and POS-Tagging (i.e. Part-of-Speech Tagging) for the process of transliteration. Stemming refers to the removal of suffixes from the root word. Root word is actually the basic word to which suffixes get added. For example, in સ્ત્રીઓમાંથી (striiomaaNthii) the root is સ્ત્રી and the suffix is ઓમાંથી.These modules prove to be beneficial in the Natural Language Processing environment for morphologically rich languages.

The rest of the paper is arranged as follows: Section 2 describes the previous history of the related work which is followed by Section3 which describes the proposed work. Evaluation and Results have been focused on in Section 4. Finally we conclude the paper with some enhancements for future work in Section 5.

## 2. LITERATURE REVIEW

Stemming was actually introduced by Lovins [1] who in 1968 proposed the use of it in Natural Language Processing applications. Two more stemming algorithms were proposed by Hafer and Weiss [2] and Paice [3]. Martin Porter [4] in 1980 suggested a suffix stripping algorithm which is still considered to be a standard stemming algorithm. Another approach to stemming was proposed by Frakes and Baeza- Yates [5] who proposed the use of term indexes and its root word in a table lookup. With the improvement in processing capabilities, there was a paradigm shift from purely rule-based techniques to statistical/ machine learning approaches. Goldsmith [6][7] proposed an unsupervised approach to model morphological variants of European languages. Snover and Brent [8] proposed a Bayesian model for stemming of English and French languages. Freitag [9] proposed an algorithm for clustering of words using co-occurrence information. For Indian languages, Larkey *et al.* [10] used 27 rules to implement a stemmer for Hindi. Ramanathan and Rao [11] used the same approach, but used some more rules for stemming. Dasgupta and Ng [12] proposed an unsupervised morphological stemmer for Bengali. Majumder *et al.* [13] proposed a cluster based approach based on string distance measures which required no linguistic knowledge. Pandey and Siddiqui [14] proposed an unsupervised approach to stemming for Hindi, which was mostly based on the work of Goldsmith.

Considering the research work for part-of-speech tagging, Church [15] proposed n-gram model for tagging, which was then extended as HMM by Cutting *et al.* [16] in 1992. Brill [17] proposed a tagger based on transformation-based learning. Ratnaparkhi [18] proposed Maximum Entropy algorithm. Many researchers have recently proposed taggers with different approaches. Ray *et al.* [19] have proposed a morphology-based disambiguation for Hindi POS tagging. Dalal *et al.* [20] have proposed Feature Rich POS Tagger for Hindi. Patel and Gali [21] have proposed a tagging scheme for Gujarati using Conditional Random Fields. A rule-based Tamil POS-Tagger was developed by Arulmozhi *et al.* [22]. Arulmozhi and Sobha [23] have developed a hybrid POS-Tagger for relatively free word order language. Similarly for Bangla, Chowdhury *et al.* [24] and Sediqqui *et al.* [25] have done significant research in the area of POS-Tagging. Antony and Soman [26] used kernel-based approach for Kannada POS-Tagging. Again a paradigm shift has been observed from purely rule-based schemes to statistical techniques. Taggers for many Indian languages have been proposed but still more work needs to be done as compared to European languages.

Moving towards the work for transliteration, Kirschenbaum and Wintner [27] have proposed a lightly supervised transliteration scheme. Arababi *et al*. [28] used a combination of neural net and expert systems for transliteration. Praneeth *et al*. [29] at LTRC, IIIT-H proposed a language-independent schema using character aligned models. Malik *et al*. [30] followed a hybrid approach for Urdu-Hindi transliteration. Joshi and Mathur [31] proposed the use of phonetic mapping based English-Hindi transliteration system which created a mapping table and a set of rules for transliteration of text. Joshi *et al*. [32] also proposed a predictive approach of for English-Hindi transliteration where the authors provided a suggestive list of possible text that the user entered. They looked at the partial text and tried to provide possible complete list as the suggestive list that the user could accept or provide their own input text. The use of transliteration has been proposed by many researchers for natural language processing and information retrieval applications.

## 3. PROPOSED WORK

Gujarati is a highly inflectional language as stated earlier. It has a free word-order. There are three genders in Gujarati- Feminine, Masculine and Neuter/Neutral. Suffixes get added to the stems giving the various morphological variants of the same root word.

We propose the use of stemming and POS-Tagging for the purpose of transliteration. Figure 1 shows our system.

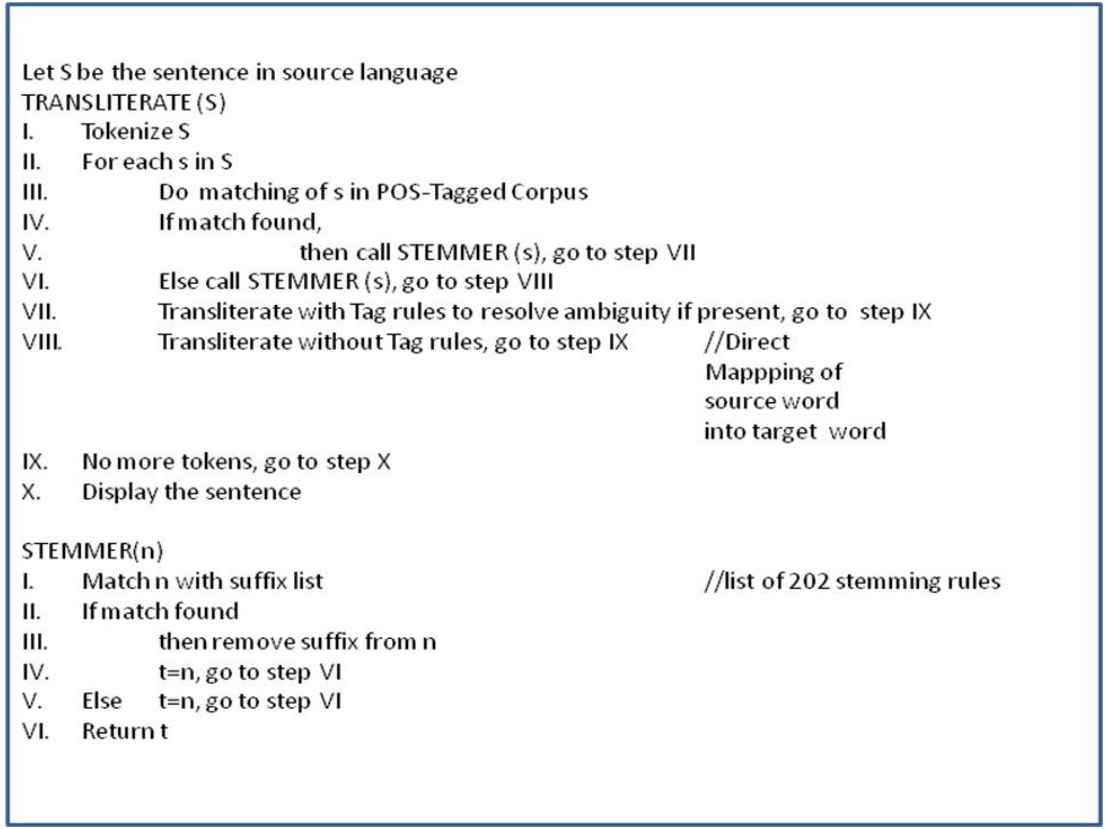

Figure1. Transliteration assisted with stemming and part-of-speech tagging

Many ambiguities are observed while we design a translation model from Gujarati-Hindi. One such ambiguity is differentiation of the suffix ે in different cases. Suppose we have the sentence

રામે મને રિપોર્ટ આપી.  →  राम ने मुझे रिपोर्ट दी|
(Raame mane riport aapii.)      (Raam ne mujhe riport dii.)
(Meaning: Ram gave me the report.)

મારા ઘરે એક બિલાડી છે.  →  मेरे घर पर एक बिल्ली है|
(Maaraa ghare ek bilaadii chhe)  (Mere ghar par ek billi hai)
(Meaning: There is a cat at my home.)

If these two sentences are carefully observed, the suffix serves different purpose. Hence it is the tag that makes a difference here. રામે is a proper noun and ઘરે is a locative noun. Hence to differentiate if a tagged corpus is applied, then during translation if the meanings are not available in the corpus and only the tags are available then the transliterated text will be the actual translation. Similarly, the suffix ીએ poses an ambiguity.

| ચાલો ઘેર ચાલીએ. | → | चलो घर चलें| |
| (Chaalo gher chaaliie.) | | (Chalo ghar chaleN.) |
| (Meaning: Let us go home.) | | |
| રશ્મીએ કિતાબ આપી. | → | रश्मी ने किताब दी| |
| (Rashmiie kitaab aapii.) | | (Rashmii ne kitaab dii.) |
| (Meaning: Rashmi gave the book.) | | |

ચાલીએ is a verb whereas, રશ્મીએ is a proper noun.

We created a raw corpus of 5400 POS-tagged sentences and used 202 stemming and tagging rules to assist transliteration. The POS-Tagged corpus is a collection of text files having the sentences in the source language in the form- word_part-of-speech, e.g. प्रतिबंध_NN. The strings in the source language are first checked in the tagged corpus so that the word class can be obtained and then stemming is applied which ensures the extraction of the correct root. Transliteration is hence first refined by these modules. So whenever there is an ambiguity in suffixes (i.e. stemming process), corresponding tags resolve the problem of transliteration. These modules can hence help in ambiguity resolution If the corresponding tag is not found in the tagged corpus, naive transliteration is done where direct mapping from the source language into the target one is applied.

## 4. EVALUATION AND RESULTS

We tested our system on a total of 500 Sentences. The observed results are as follows:

| | |
|---|---|
| Total number of Sentences tested | 500 |
| Total number of words tested | 7500 |
| Words for which transliteration and translation are same | 4086 |
| Percentage of words for which translation and transliteration are same | 54.48 |
| Number of words for which transliteration was wrong | 518 |
| Percentage of words for which transliteration was wrong | 6.91 |
| Percentage efficiency of transliteration | 93.09 |

Table 1.Table showing evaluated results

Hence for 54.48% of Gujarati words translation and transliteration are same. The efficiency of our transliteration scheme is 93.09% (about 90%).

## 5. CONCLUSION AND FUTURE WORK

We followed a hybrid approach – a mix of rule-based and corpus-based approach, where we used POS-Tagged corpus and stemming rules to assist the process of transliteration. We achieved 93.09% overall efficiency of the transliteration scheme which makes it a promising approach. It was observed that 54.48% of the Gujarati words have the same translation and transliteration. Such a scheme not only reduces length of the corpus for the translation model but also it helps in ambiguity resolution. It can be used for other morphologically rich Indian languages as well. As an immediate extension to this work, we plan further to include machine learning approaches and focus on each and every aspect of the scheme so that more accuracy in the transliteration process can be achieved.

## AUTHORS

**Juhi Ameta** has completed her M.tech. in Computer Science from Banasthali Vidyapith, Rajasthan and is a Gold-medalist of her batch. She is currently working as an Assistant Professor at Cummins College of Engineering for Women, Pune, India. Her research interests include Natural Language Processing and Machine Translation. She has worked on EILMT Project funded by DIT, Govt. of India. Her research paper entitled "A Lightweight Stemmer for Gujarati" was published by CSI, Annual National Conference, Ahmedabad Chapter, December 2011.

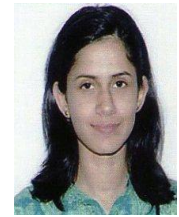

**Nisheeth Joshi** is a researcher working in the area of Machine Translation. He has been primarily working in design and development of evaluation Matrices in Indian languages. Besides this he is also actively involved in the development of MT engines for English to Indian Languages. He is one of the expert empanelled with TDIL programme, Department of electronics Information Technology (DEITY), Govt. of India, a premier organization which foresees Language Technology Funding and Research in India. He has several publications in various journals and conferences and also serves on the Programme Committees and Editorial Boards of several conferences and journals.

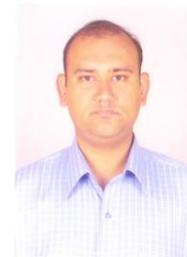

**Iti Mathur** is an Assistant Professor at Banasthali University, India. Her research interests are in the area of Ontological engineering, soft computing, machine translation, and information retrieval. She is also a Co-Principal Investigator of English to Indian Language Machine Translation Development System Funded by Govt. of India. The project is a consortium mode project, where 13 institutions are developing machine translators from English to 8 different Indian languages. She has published several papers in natural language processing and information retrieval. She is also Editorial Board Member/Program Committee Member and reviewer of various journals and conferences. She is a Member of IEEE, USA, ACM, USA and CSI, India.

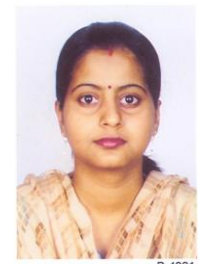